\def\year{2020}\relax
\documentclass[letterpaper]{article} 
\usepackage{aaai20}  
\usepackage{times}  
\usepackage{helvet} 
\usepackage{courier}  
\usepackage[hyphens]{url}  
\usepackage{graphicx} 
\usepackage{graphicx}  
\usepackage{bm}
\usepackage{amsmath,amssymb,amsfonts}
\usepackage{algorithmic}
\usepackage{textcomp}
\usepackage{xcolor}
\usepackage{multirow}
\usepackage{booktabs}
\usepackage{threeparttable}
\usepackage{tabularx}
\usepackage{algorithm}
\usepackage{easylist}

\DeclareMathOperator*{\argmax}{arg\,max}
\usepackage{amssymb}
\usepackage{pifont}
\newcommand{\cmark}{\ding{51}}%
\newcommand{\xmark}{\ding{55}}%
\title{Unsupervised Domain Adaptation via Structured Prediction Based Selective Pseudo-Labeling}
\author{Qian Wang, \textsuperscript{\rm 1}  Toby P. Breckon \textsuperscript{\rm 1,2}\\ 
\textsuperscript{\rm 1} Department of Computer Science, Durham University, United Kingdom\\
\textsuperscript{\rm 2} Department of Engineering, Durham University, United Kingdom\\
qian.wang173@hotmail.com, toby.breckon@durham.ac.uk 
}
\begin{document}

\maketitle

\begin{abstract}
Unsupervised domain adaptation aims to address the problem of classifying unlabeled samples from the target domain whilst labeled samples are only available from the source domain and the data distributions are different in these two domains. As a result, classifiers trained from labeled samples in the source domain suffer from significant performance drop when directly applied to the samples from the target domain. 
To address this issue, different approaches have been proposed to learn domain-invariant features or domain-specific classifiers. In either case, the lack of labeled samples in the target domain can be an issue which is usually overcome by pseudo-labeling. Inaccurate pseudo-labeling, however, could result in catastrophic error accumulation during learning. In this paper, we propose a novel selective pseudo-labeling strategy based on structured prediction. The idea of structured prediction is inspired by the fact that samples in the target domain are well clustered within the deep feature space so that unsupervised clustering analysis can be used to facilitate accurate pseudo-labeling. Experimental results on four datasets (i.e. Office-Caltech, Office31, ImageCLEF-DA and Office-Home) validate our approach outperforms contemporary state-of-the-art methods. 
\end{abstract}

\section{Introduction}\label{sec:introduction}
Domain adaptation problems exist in many real-world applications. Unsupervised Domain Adaptation (UDA) aims to address problems where the unlabeled test samples and the labeled training samples are from different domains (dubbed target and source domains, respectively). Models learned with labeled samples from the source domain suffer from significant performance reduction when they are directly applied to the target samples without domain adaptation. Such reduced performance is mainly due to the domain shift characterized by the difference of data distributions between the two domains. To address this issue,  approaches to UDA have been proposed trying to align the source and target domains by learning a joint subspace so that samples from either domain can be projected into this common subspace. Different algorithms were employed to promote the separability of target samples in such subspaces. The use of visual features extracted by deep models pre-trained on the large-scale ImageNet dataset \cite{deng2009imagenet} further facilitates these feature transformation based approaches. On the other hand, deep learning models are used to learn domain invariant features in an end-to-end manner. The gradient reversal layer \cite{ganin2015unsupervised} and adversarial learning \cite{tzeng2017adversarial} have been employed for this purpose.

In the UDA problem, both labeled samples from the source domain and unlabeled samples from the target domain are assumed to be available during model training and hence it is a transductive learning problem. The opportunity of transductive learning is when the algorithm can explore the test data (samples from the target domain in the case of UDA). One effective method is to assign pseudo-labels to the target samples so that samples from both domains can be combined and ready for supervised learning. The accuracy of pseudo-labeling plays an important role in the whole learning process. In most existing methods, the target samples are pseudo-labeled independently once the classifier is trained overlooking the structural information underlying the target domain. 

In this paper, we explore such structural information via unsupervised learning (i.e. $K$-means) and propose a novel UDA approach based on selective pseudo-labeling and structured prediction. Specifically, our approach tries to learn a domain invariant subspace by Supervised Locality Preserving Projection (SLPP) using both labeled source data and pseudo-labeled target data. The accuracy of pseudo-labeling is promoted by structured prediction and progressively selections.  The contributions of this work can be summarized as follows:

\begin{itemize}
\item[--]{a novel iterative learning algorithm is proposed for UDA using SLPP based subspace learning and the selective pseudo-labeling strategy.}
\item[--]{structured prediction is employed to explore the structural information within the target domain to promote the accuracy of pseudo-labeling and domain alignment.}
\item[--]{thorough comparative experiments and ablation studies are conducted to demonstrate the proposed approach can achieve new state-of-the-art performance on four benchmark datasets.}
\end{itemize}


\section{Related Work}\label{sec:relatedwork}
Early works on UDA problems aim to align the marginal distributions of source and target domains \cite{gong2012geodesic,ganin2015unsupervised,sun2016return}. With aligned marginal distributions, it is not guaranteed to produce good classification results as the conditional distribution of the target domain can be misaligned with that of the source domain. This is mainly due to the lack of labeled target samples.  To overcome this issue, many UDA approaches have employed pseudo-labeling strategies during learning \cite{long2013transfer,zhang2017joint,wang2018visual,pei2018multi,zhang2018collaborative,wang2019unifying,chen2019progressive}. Pseudo labeling the target samples allows to align the conditional distributions of source and target domains with traditional supervised learning algorithms. To give a sketch of how existing works handle the issue of lacking labeled samples in the target domain, in this section, we review related works on UDA by dividing them into three categories: \textit{approaches without pseudo-labeling}, \textit{pseudo-labeling without selection} and \textit{pseudo-labeling with selection}. 

\subsection{Approaches without Pseudo-Labeling}\label{sec:softlabeling} 
Approaches to UDA aiming to align the marginal distributions of source and target domains can be realized via minimising the Maximum Mean Discrepancy (MMD) \cite{long2014transfer,sun2016return}. The same idea has also been employed in deep learning based approaches to learning domain invariant features \cite{long2015learning,sun2016deep,long2016unsupervised,chen2018joint}. Alternatively, the same goal can be achieved by the gradient reversal layer \cite{ganin2015unsupervised,ganin2016domain} or generative adversarial loss \cite{tzeng2017adversarial}. Although these models can learn domain invariant features which are also discriminative for the source domain, the separability of target samples is not guaranteed since the conditional distributions are not explicitly aligned. More recently, domain-symmetric networks were proposed to promote the alignment of joint distributions of feature and category across source and target domains \cite{zhang2019domain}. In contrast to these approaches, pseudo-labeling target samples is another effective way to promote the alignment of conditional distributions.

\subsection{Pseudo-Labeling without Selection}\label{sec:withoutselection}
Pseudo-labeling without selection assigns pseudo-labels to all samples in the target domain. Two strategies, i.e. hard labeling \cite{long2013transfer,zhang2017joint,wang2018visual} and soft labeling \cite{pei2018multi}, have been employed in existing works.
The strategy of hard labeling assigns a pseudo-label $\hat{y}$ to each unlabeled sample without considering the confidence. It can be achieved by a classifier trained on labeled source samples \cite{long2013transfer,zhang2017joint,wang2018visual}. The pseudo-labeled target samples together with labeled source samples are used to learn an improved classification model.  By iterative learning, the pseudo-labeling is expected to be progressively more accurate until convergence. The problem of such hard pseudo-labeling is that mis-labeled samples by a weak classifier in the initial stage of the iterative learning can cause serious harm to the subsequent learning process. 
To address this issue, soft labeling was employed in \cite{pei2018multi}. The strategy of soft labeling assigns the conditional probability of each class $p(c|x)$ given a target sample $x$, which results in a pseudo-labeling vector $\hat{y} = \{p(c_1|x),p(c_2|x),...,p(c_{|\mathcal{C}|}|x)\} \in [0,1]^{|\mathcal{C}|}$, where $|\mathcal{C}|$ is the number of classes. The soft label $\hat{y}$ can be updated during iterative learning and the final classification results can be derived by selecting the class with the highest probability. Soft labeling is naturally suitable for neural network based approaches whose outputs are usually a vector of conditional probabilities. For instance, in the Multi-Adversarial Domain Adaptation (MADA) approach \cite{pei2018multi}, the soft pseudo-label of a target sample is used to determine how much this sample should be attended to different class-specific domain discriminators. 

\subsection{Pseudo-Labeling with Selection} \label{sec:withselection}
Selective pseudo-labeling is the other way to alleviate the mis-labeling issue \cite{zhang2018collaborative,wang2019unifying,chen2019progressive}. Similar to the soft labeling strategy, selective pseudo-labeling also takes into consideration the confidence in target sample labeling but in a different manner. Specifically, a subset of target samples are selected to be assigned with pseudo labels and only these pseudo-labeled target samples are combined with source samples in the next iteration of learning. The idea is that at the beginning the classifier is weak so that only a small fraction of the target samples can be correctly classified. When the classifier gets stronger after each iteration of learning, more target samples can be correctly classified hence should be pseudo-labeled and participate in the learning process. One key factor in such algorithms is the criterion of sample selection for pseudo-labeling. An easy-to-hard strategy was employed in \cite{chen2019progressive}. Target samples whose similarity scores are higher than a threshold are selected for pseudo-labeling and this threshold is updated after each iteration of learning so that more unlabeled target samples can be selected. One limitation of this sample selection strategy is the risk of biasing to ``easy" classes and the selected samples in the first iterations can be dominated by these ``easy" classes. As a result, the learned model will be seriously biased to the ``easy" classes. To address this issue, a class-wise sample selection strategy was proposed in \cite{wang2019unifying}. Samples are selected for each class independently so that pseudo-labeled target samples will contribute to the alignment of conditional distribution for each class during learning.
In this paper, we propose a novel approach to selective pseudo-labeling by exploring the structural information within the unlabeled target samples.

\section{Proposed Method}\label{sec:method}
\begin{figure}
	\centering
	{\includegraphics[width=0.47\textwidth]{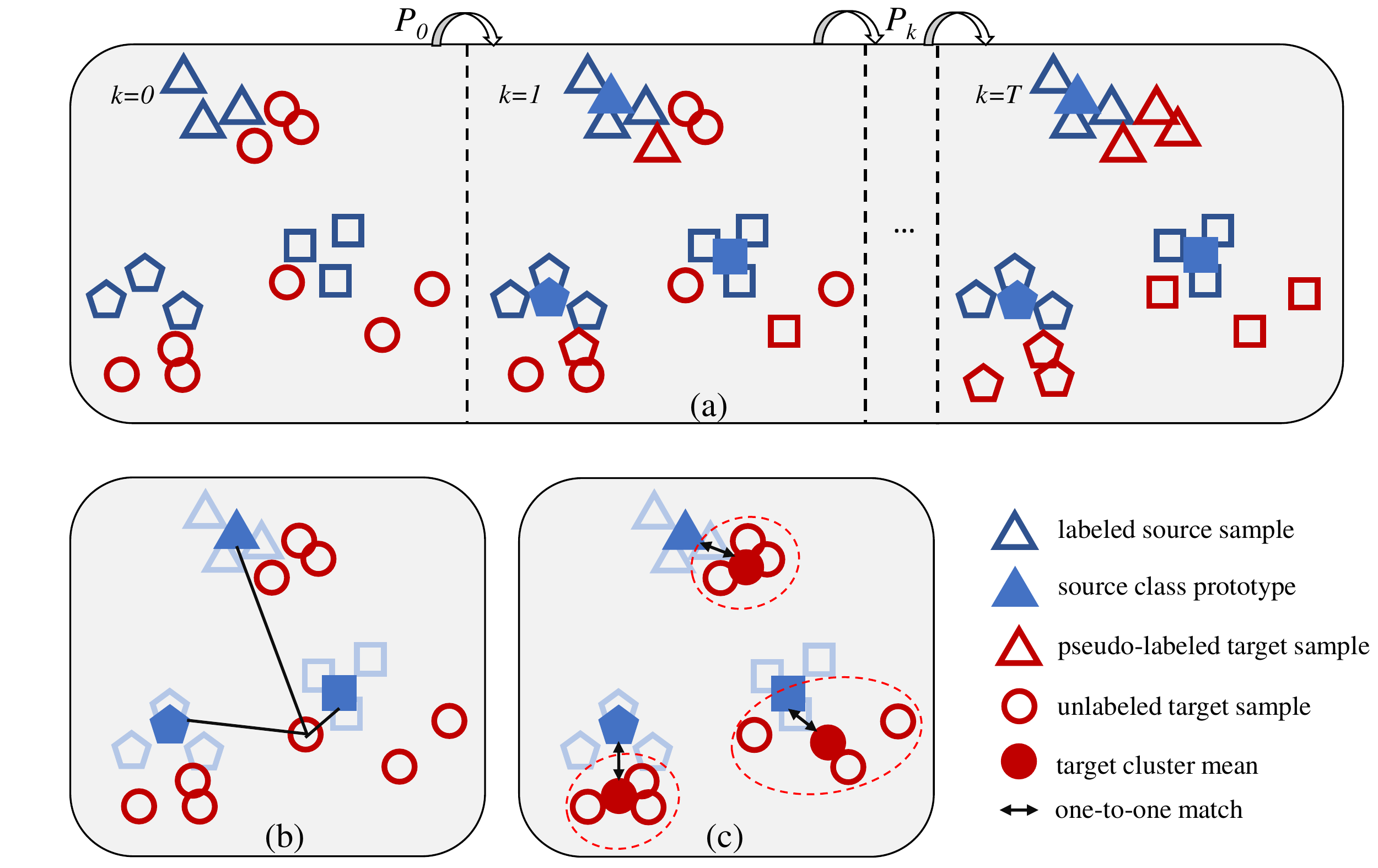}}
	{\caption{The framework of our proposed approach. (a) iterative learning process (b) pseudo-labeling via nearest class prototype (c) pseudo-labeling via structured prediction. The red and the blue colors are used for the target and source domains respectively. }
		\label{fig:framework}}
\end{figure}

The proposed method aims to align the conditional distributions of source and target domains. We employ Supervised Locality Preserving Projection (SLPP) \cite{he2004locality} as an enabling technique to learn a projection matrix $\mathbf{P}$ which maps samples from both domains into the same latent subspace. The learned subspace is expected to have favourable properties that projections of samples from the same class will be close to each other regardless of which domain they are from. In the subspace, a classifier such as nearest neighbour is used to classify unlabeled target samples. By combining both pseudo-labeled target samples and labeled source samples, the projection matrix $\mathbf{P}$ can be updated. As a result, an iterative learning process is employed to improve the projection learning and pseudo-labeling alternately as illustrated in Figure \ref{fig:framework} (a). 

Following \cite{wang2019unifying}, we use the nearest class prototype (NCP) for pseudo-labeling. Source class prototypes are computed by averaging projections of source samples from the same class in the same space. Target samples can be pseudo-labeled by measuring the distances to these class prototypes (see Figure \ref{fig:framework}(b)). This method overlooks the intrinsic structural information in the target domain, resulting in sub-optimal pseudo-labeling results. Therefore, we explore the structural information underlying the target domain via clustering analysis (e.g., $K$-means). The clusters of target samples are matched with source classes via structured prediction \cite{zhang2016zero,wang2017zero} so that target samples can be labeled collectively according to which cluster they belong to (see Figure \ref{fig:framework} (c)). We calculate the distances of target samples to cluster centers as the criterion for selective pseudo-labeling. The samples close to the cluster center are more likely to be selected for pseudo-labeling and participate in the projection learning in the next iteration of learning. In contrast to the existing sample selection strategies \cite{chen2019progressive,wang2019unifying}, the structured prediction based method tends to select samples far away from the source samples which enables a faster domain alignment. Moreover, since these two methods are intrinsically different from each other, a combination of two is expected to further benefit the learning process.

In the following subsections, we give the formulation of the problem, describe each component of the proposed method in detail and analyse the complexity of the algorithm.

\subsection{Problem Formulation}\label{sec:problem}
Given a labeled dataset $\mathcal{D}^s = \{(\bm{x}^s_i,y^s_i)\}, i = 1,2,...,n_s$ from the source domain $\mathcal{S}$, $\bm{x}^s_i \in \mathbb{R}^{d}$ represents the feature vector of $i$-th labeled sample in the source domain, $d$ is the feature dimension and $y^s_i \in \mathcal{Y}^s$ denotes the corresponding label. UDA aims to classify an unlabeled data set $\mathcal{D}^t = \{\bm{x}^t_i\}, i=1,2,...,n_t$ from the target domain $\mathcal{T}$, where $\bm{x}^t_i \in \mathbb{R}^{d}$ represents the feature vector in the target domain. The target label space $\mathcal{Y}^t$ is equal to the source label space $\mathcal{Y}^s$. It is assumed that both the labeled source domain data $\mathcal{D}^s$ and the unlabeled target domain data $\mathcal{D}^t$ are available for model learning.

\subsection{Dimensionality Reduction}\label{sec:preprocessing}
High dimensional features contain redundant information and thus result in unnecessary computation. We apply dimensionality reduction to the high dimensional deep features as the preprocessing. Principle Component Analysis (PCA) is selected in our work since it has been successfully used in other existing UDA approaches \cite{long2013transfer,long2014transfer}. Feature vectors of samples from source and target domains are concatenated as a matrix $\mathbf{X} = [\bm{x}_1^s,...,\bm{x}_{n_s}^s,\bm{x}_1^t,...,\bm{x}_{n_t}^t] \in \mathbb{R}^{d\times n}$, where $n=n_s+n_t$. The centering matrix is denoted as $\mathbf{H}= \mathbf{I}-\frac{1}{n} \mathbf{1}$, where $\mathbf{1}$ is an $n\times n$ matrix of ones. The objective of PCA is:
\begin{equation}\label{eq:pca}
\max_{\mathbf{V}^T\mathbf{V}=\mathbf{I}} tr(\mathbf{V}^T\mathbf{X}\mathbf{H}\mathbf{X}^T\mathbf{V})
\end{equation}
where tr($\cdot$) denotes the trace of a matrix.

The problem defined in Eq.(\ref{eq:pca}) is equivalent to the following eigenvalue problem:
\begin{equation}\label{eq:pca_eig}
\mathbf{X}\mathbf{H}\mathbf{X}^T \bm{v} = \phi \bm{v}.
\end{equation}
By solving the eigenvalue problem, we can have the PCA projection matrix $\mathbf{V} = [\bm{v}_1,...,\bm{v}_{d_1}] \in \mathbb{R}^{d\times d_1}$ and the lower-dimensional features:
\begin{equation}\label{eq:pca_proj}
    \mathbf{\tilde{X}} = \mathbf{V}^T \mathbf{X} 
\end{equation}
where $\mathbf{\tilde{X}} \in \mathbb{R}^{d_1\times n}$ and $d_1 \leq d$ is the dimensionality of the feature space after applying PCA.

Since PCA is a linear feature transformation, $L$2 normalization is applied to each feature vector in $\mathbf{\tilde{X}}$ as $\bm{\tilde{x}} \leftarrow \bm{\tilde{x}}/||\bm{\tilde{x}}||_2$.
The use of $L$2 normalization forces samples of both source and target domains distributed on the surface of the same hyper-sphere which helps to align data from different domains \cite{wang2017zero}. Our experimental results in this study also provide empirical evidence that such sample normalization is beneficial to superior performance. 

\subsection{Domain Alignment}\label{sec:method_da}
The lower-dimensional feature space $\mathcal{\tilde{X}}$ learned by PCA and sample normalisation exhibit favourable properties of domain alignment. However, it is learned in an unsupervised manner and is thus not sufficiently discriminative. To promote the class-wise alignment of two domains, we use the supervised locality preserving projection \cite{he2004locality} as an enabling technique to learn a domain invariant yet discriminative subspace $\mathcal{Z}$ from $\mathcal{\tilde{X}}$.


The objective of SLPP is to learn a projection matrix $\mathbf{P}$ by minimizing the following cost function :
\begin{equation}
\label{eq:cost}
\min_{\mathbf{P}} \sum_{i,j} || \mathbf{P}^T \bm{\tilde{x}_i} - \mathbf{P}^T \bm{\tilde{x}_j}||_2^2 \mathbf{M}_{ij},
\end{equation}
where $\mathbf{P} \in \mathbb{R}^{d_1\times d_2}$ and $d_2 \leq d_1$ is the dimensionality of the learned space; $\bm{\tilde{x}}_i$ is the $i$-th column of the labeled data matrix $\mathbf{\tilde{X}}^l\in \mathbb{R}^{d_1 \times (n_s+n_t')}$ and $\mathbf{\tilde{X}}^l$ is a collection of $n_s$ labeled source data and $n_t'$ selected pseudo-labeled target data. The similarity matrix $\mathbf{M} \in \mathbb{R}^{(n_s+n_t')\times(n_s+n_t')}$ is defined as follows:
\begin{equation}
\label{eq:sim}
\mathbf{M}_{ij} = \left \{
\begin{array}{ll}
1, & y_i = y_j,\\
0 ,&  otherwise.
\end{array}
\right.
\end{equation}
The idea is that samples from the same class should be projected close to each other in the subspace regardless of which domain they are originally from. Eq (\ref{eq:sim}) is a simplified version of MMD matrices used in \cite{long2013transfer,zhang2017joint,wang2018visual} where the domain differentiation is reserved while we try to promote the domain invariance. Our definition of similarity matrix in Eq (\ref{eq:sim}) also differs from that in the original LPP formulation where local structure of the samples is considered by defining the similarity value $\mathbf{M}_{ij}$ based on the distance between $\bm{\tilde{x}_i}$ and $\bm{\tilde{x}_j}$.

Following the treatment in \cite{he2004locality,wang2017zero}, the objective can be rewritten as:
\begin{equation}
\label{eq:obj}
\max_{\mathbf{P}} \frac{tr(\mathbf{P}^T\mathbf{\tilde{X}}^l\mathbf{D} \mathbf{\tilde{X}}^{lT} \mathbf{P})}{tr(\mathbf{P}^T(\mathbf{\tilde{X}}^l\mathbf{L} \mathbf{\tilde{X}}^{lT} + \mathbf{I})\mathbf{P})}
\end{equation}
where $\mathbf{L}=\mathbf{D}-\mathbf{M}$ is the laplacian matrix, $\mathbf{D}$ is a diagonal matrix with $\mathbf{D}_{ii}=\sum_j \mathbf{M}_{ij}$ and the regularization term $tr(\mathbf{P}^T \mathbf{P})$ is added for penalizing extreme values in the projection matrix $\mathbf{P}$.

The problem defined in Eq.(\ref{eq:obj}) is equivalent to the following generalized eigenvalue problem:
\begin{equation}
\label{eq:eig}
\mathbf{\tilde{X}}^l\mathbf{D} \mathbf{\tilde{X}}^{lT}\bm{p} = \lambda (\mathbf{\tilde{X}}^l\mathbf{L} \mathbf{\tilde{X}}^{lT} + \mathbf{I}) \bm{p},
\end{equation}
solving the generalized eigenvalue problem gives the optimal solution $\mathbf{P}=[\bm{p}_1, ..., \bm{p}_{d_2}]$ where $\bm{p}_1,...,\bm{p}_{d_2}$ are the eigenvectors corresponding to the largest $d_2$ eigenvalues.

Learning the projection matrix $\mathbf{P}$ for domain alignment requires labeled samples from both source and target domains. To get pseudo-labels of target samples for projection learning, we describe pseudo-labeling methods via nearest class prototype and structured prediction respectively in the following sub-sections. 

\subsection{Pseudo-Labeling via Nearest Class Prototype (NCP)}\label{sec:method_plnn}
Unlabeled target samples can be labeled in the learned subspace $\mathcal{Z}$ where the projections of source and target samples are computed by:
\begin{equation}\label{eq:slpp_proj}
\begin{array}{cc}
     \bm{z}^s = \mathbf{P}^T \bm{\tilde{x}}^s, \quad
     \bm{z}^t = \mathbf{P}^T \bm{\tilde{x}}^t.
\end{array}
\end{equation}
We sequentially apply centralisation (i.e. mean subtraction, $\bm{z} \leftarrow \bm{z}-\bar{\bm{z}}$, where $\bar{\bm{z}}$ is the mean of all source and target sample projections) and $L$2 normalisation to $\bm{z}$ to promote the separability of different classes in the space $\mathcal{Z}$.

The class prototype for class $y\in \mathcal{Y}$ is defined as the mean vector of the projected source samples whose labels are $y$, which can be computed by:
\begin{equation}\label{eq:prototype}
\bm{\bar{z}}_y^s = \frac{\sum_{i=1}^{n_s} \bm{z}_i^s \delta(y,y_i^s)}{\sum_{i=1}^{n_s}\delta(y,y_i^s)},
\end{equation}
where $\delta(y,y_i)=1$ if $y=y_i$ and 0 otherwise. After applying $L$2 normalization to the class prototypes $\bm{\bar{z}}_y^s, y = 1,..., |\mathcal{Y}|$, where $|\mathcal{Y}|$ denotes the number of classes, we can derive the conditional probability of a given target sample $\bm{x}^t$ belonging to class $y$:
\begin{equation}\label{eq:conditional}
p_1(y|\bm{x}^t) = \frac{\exp{(-||\bm{z}^t-\bm{\bar{z}}_y^s||)}}{\sum_{y=1}^{|\mathcal{Y}|}\exp{(-||\bm{z}^t-\bm{\bar{z}}_y^s||)}}.
\end{equation}

\subsection{Pseudo-Labeling via Structured Prediction (SP)}\label{sec:method_plsp}
Pseudo-labeling via nearest class prototype does not consider the intrinsic structure of the target samples which provides useful information for target samples classification. To explore such structure information, we employ structured prediction for pseudo-labeling. Specifically, we use $K$-means to generate $|\mathcal{Y}|$ clusters over projection vectors $\bm{z}^t$ of all target samples. The cluster centers are initialised with class prototypes calculated by Eq. (\ref{eq:prototype}). Subsequently, we establish a one-to-one match between a cluster from the target domain and a class from the source domain so that the sum of distances of all the matched pairs of the cluster center and the class prototype is minimised. Let $\mathbf{A} \in \{0,1\}^{|\mathcal{Y}|\times |\mathcal{Y}|}$ denote the one-to-one matching matrix where $\mathbf{A}_{ij} = 1$ indicates that the $i$-th target cluster is matched with the $j$-th source class. The optimisation problem can be formulated as follows:
\begin{equation}\label{eq:matching}
\begin{array}{cc}
\min\limits_{\mathbf{A}} \sum_{i=1}^{|\mathcal{Y}|} \sum_{j=1}^{|\mathcal{Y}|} \mathbf{A}_{ij} d(\bm{\bar{z}}_i^t,\bm{\bar{z}}_j^s)\\
\\
s.t. \quad \forall i,\sum\limits_j\mathbf{A}_{ij}=1; \forall j, \sum\limits_i \mathbf{A}_{ij} = 1,
\end{array}
\end{equation}
where $\bm{\bar{z}}_i^t$ denotes the $i$-th cluster center in the target domain. This problem can be efficiently solved by linear programming according to \cite{zhang2016zero,wang2017zero}.

Let $\bm{\bar{z}}_y^t$ denote the cluster center corresponding to the class $y$, similar to Eq. (\ref{eq:conditional}), we can calculate the conditional probability of a given target sample $\bm{x}^t$ belonging to class $y$:
\begin{equation}\label{eq:conditional2}
p_2(y|\bm{x}^t) = \frac{\exp{(-||\bm{z}^t-\bm{\bar{z}}_y^t||)}}{\sum_{y=1}^{|\mathcal{Y}|}\exp{(-||\bm{z}^t-\bm{\bar{z}}_y^t||)}}.
\end{equation}

\subsection{Iterative Learning with Selective Pseudo-Labeling (SPL)}\label{sec:method_learning}
We use an iterative learning strategy to learn the projection matrix $\mathbf{P}$ for domain alignment and improved pseudo-labeling for target samples alternately. Although either of the two pseudo-labeling methods described above is able to provide useful pseudo-labeled target samples for projection learning in the next iteration, they are intrinsically different. Pseudo-labeling via nearest class prototype tends to output high probability to the samples close to the source data, whilst structured prediction is confident in samples close to the cluster center in the target domain regardless how far they are from the source domain. We advocate to take advantage of the complementarity of these two methods via a simple combination of Eq.(\ref{eq:conditional}) and Eq.(\ref{eq:conditional2}) as follows:
\begin{equation}\label{eq:probfusion}
p(y|\bm{x}^t) = \max \{p_1(y|\bm{x}^t), p_2(y|\bm{x}^t)\}.
\end{equation}
As a result, the pseudo-label of a given target sample $\bm{x}^t$ can be predicted by:
\begin{equation}\label{eq:pseudolabel}
\hat{y}^t = \argmax\limits_{y\in\mathcal{Y}} p(y|\bm{x}^t).
\end{equation}
Now we have pseudo labels for all the target samples as well as the probability of these pseudo labels, denoted as a set of triplets $\mathcal{\hat{D}}^t = \{(\bm{x}_i^t,\hat{y}_i^t,p(\hat{y}_i^t|\bm{x}_i^t))\},i=1,...,n_t$ .

Instead of using all the pseudo-labeled target samples for the projection learning, we progressively select a subset $\mathcal{S}_k \subseteq \mathcal{\hat{D}}^t$ containing $kn_t/T$ target samples in the $k$-th iteration, where $T$ is the number of iterations of the learning process. One straight forward strategy is to select top $kn_t/T$ samples with highest probabilities from $\mathcal{\hat{D}}^t$. However, this strategy has a risk of only selecting samples from specific classes while overlooking the other classes. To avoid this, we do the class-wise selection so that target samples pseudo-labeled as each class have the same opportunity to be selected. Specifically, for each class $c \in \mathcal{Y}$, we first pick out $n_t^c$ target samples pseudo-labeled as class $c$ from which we select top $kn_t^c/T$ high-probability samples to form $\mathcal{S}_k$.
The overall algorithm is summarized in Algorithm \ref{alg:uda}.
\begin{algorithm}[tb]
	\caption{Unsupervised Domain Adaptation Using Selective Pseudo-Labeling}
	\label{alg:uda}
	\renewcommand{\algorithmicrequire}{\textbf{Input:}}
	\renewcommand{\algorithmicensure}{\textbf{Output:}}
	\begin{algorithmic}[1]
		\REQUIRE Labeled source data set $\mathcal{D}^s = \{(\bm{x}^s_i,y^s_i)\}, i = 1,2,...,n_s$ and unlabeled target data set $\mathcal{D}^{t}=\{\bm{x}_i^t\},i=1,2,...,n_{t}$, dimensionality of PCA and SLPP subspace $d_1$ and $d_2$, number of iteration $T$.
		\ENSURE The projection matrix $\mathbf{P}$ and predicted labels $\{\hat{y}^t\}$ for target samples.
		\STATE Initialize $k=0$;
		\STATE Dimensionality reduction by Eq. (\ref{eq:pca_proj});
		\STATE Learn the projection $\mathbf{P}_0$ using only source data $\mathcal{D}^s$;
		\STATE Assign pseudo labels for all target data using Eq. (\ref{eq:pseudolabel});
		\WHILE {$k < T$}
		\STATE $k \leftarrow k+1$;
		\STATE Select a subset of pseudo-labeled target data $\mathcal{S}_k \in \mathcal{\hat{D}}^t $;
		\STATE Learn $P_k$ using $\mathcal{D}^s$ and $\mathcal{S}_k$;
		\STATE Update pseudo labels for all target data using Eq.(\ref{eq:pseudolabel}).
		\ENDWHILE
	\end{algorithmic}
\end{algorithm}

\subsection{Computational Complexity} \label{sec:method_complexity}
We analyse the computation complexity of our learning algorithm. The complexity of PCA is $\mathcal{O}(dn^2+d^3)$. The complexity of SLPP is $\mathcal{O}(2d_1n^2+d_1^3)$ which is repeated for $T$ times and leads to approximately $\mathcal{O}(T(2d_1n^2+d_1^3))$. Since the iterative learning contributes the most to the computation cost, a small value of $d_1 < d$ can make the learning process more efficient when $n$ is not too big.

\section{Experiments and Results}\label{sec:experiments}

\begin{table*}[!t]
	\centering
	{
		\centering
		\caption[]{Classification Accuracy (\%) on Office-Caltech dataset using Decaf6 features. Each column displays the results of a pair of source $\to$ target setting.}
		\label{table:uda_o10}
		\resizebox{2\columnwidth}{!}{%
			\begin{tabular}{cccccccccccccc}
				\hline
				Method & C$\to$A & C$\to$W & C$\to$D & A$\to$C&A$\to$W & A$\to$D&W$\to$C & W$\to$A & W$\to$D & D$\to$C & D$\to$A & D$\to$W & Average \\ \hline
				DDC\cite{tzeng2014deep}  & 91.9 & 85.4 & 88.8 & 85.0 & 86.1 & 89.0 & 78.0 & 84.9 & \textbf{100.0} & 81.1 & 89.5 & 98.2 & 88.2\\
				DAN\cite{long2015learning}  & 92.0 & 90.6 & 89.3 & 84.1 & \underline{91.8} & \underline{91.7} & 81.2 & 92.1 & \textbf{100.0} & 80.3 & 90.0 & 98.5 & 90.1\\
				DCORAL\cite{sun2016deep} & 92.4 & 91.1 & 91.4 & 84.7& - & - & 79.3 & - & - & 82.8 & - & - & -\\
				\hline
				CORAL\cite{sun2017correlation}& 92.0 & 80.0 & 84.7 & 83.2 & 74.6 & 84.1 & 75.5 & 81.2 & \textbf{100.0} & 76.8 & 85.5 & 99.3 & 84.7\\
				SCA\cite{ghifary2017scatter}  & 89.5 & 85.4 & 87.9 & 78.8 & 75.9 & 85.4 & 74.8 & 86.1 & \textbf{100.0} & 78.1 & 90.0 & 98.6 & 85.9 \\
				JGSA\cite{zhang2017joint} & 91.4 & 86.8 & 93.6 & 84.9 & 81.0 & 88.5 & 85.0 & 90.7 & \textbf{100.0} & 86.2 & 92.0 & \underline{99.7} & 90.0\\
				MEDA\cite{wang2018visual} & \textbf{93.4} & \textbf{95.6} & 91.1 & \textbf{87.4} & 88.1 & 88.1 & \textbf{93.2} & \textbf{99.4} & \underline{99.4} & 87.5 & \textbf{93.2} & 97.6 & \underline{92.8}\\
				CAPLS \cite{wang2019unifying} & 90.8 & 85.4 & \underline{95.5} & \underline{86.1} & 87.1 & \textbf{94.9} & \underline{88.2} & \underline{92.3} & \textbf{100.0} & \textbf{88.8} & \underline{93.0} & \textbf{100.0}& 91.8\\
				\hline
				SPL (Ours) &\underline{92.7} & \underline{93.2} & \textbf{98.7} &\textbf{87.4} & \textbf{95.3} & 89.2 & 87.0 & 92.0 & \textbf{100.0} & \underline{88.6} & 92.9 & 98.6 & \textbf{93.0}\\
				\hline
			\end{tabular}%
		}
	}
\end{table*}

\begin{table}[!t]
	\centering
	{
		\centering
		\caption[]{Classification Accuracy (\%) on Office31 dataset using either ResNet50 features or ResNet50 based deep models.
		}
		\label{table:uda_o31}
		\resizebox{1\columnwidth}{!}{%
			\begin{tabular}{cccccccc}
				\hline
				Method & \scriptsize{A$\to$W} & \scriptsize{D$\to$W} & \scriptsize{W$\to$D} & \scriptsize{A$\to$D} & \scriptsize{D$\to$A} & \scriptsize{W$\to$A} & Avg \\ \hline
				RTN\cite{long2016unsupervised} & 84.5 & 96.8 & 99.4 & 77.5 & 66.2 & 64.8 & 81.6\\
				MADA\cite{pei2018multi} & 90.0 & 97.4 & 99.6 & 87.8 & 70.3 & 66.4 & 85.2 \\
				GTA \scriptsize{\cite{sankaranarayanan2017generate}} & 89.5& 97.9 & \underline{99.8}& 87.7 & 72.8 & 71.4& 86.5\\
				iCAN\cite{zhang2018collaborative} & 92.5 & \textbf{98.8} & \textbf{100.0} & 90.1 & 72.1 & 69.9 & 87.2 \\
				CDAN-E\cite{long2018conditional} & \underline{94.1} & 98.6 & \textbf{100.0} & 92.9 & 71.0 & 69.3 & 87.7\\
				JDDA\cite{chen2018joint} & 82.6 & 95.2 & 99.7 & 79.8 & 57.4 & 66.7 & 80.2\\
				SymNets\cite{zhang2019domain} & 90.8 & \textbf{98.8} & \textbf{100.0} & \textbf{93.9} & 74.6 & 72.5 & \underline{88.4}\\
				TADA \cite{wang2019transferable} &  \textbf{94.3} & 98.7 & \underline{99.8} & 91.6 & 72.9 & 73.0 & \underline{88.4}\\
				\hline
				MEDA\cite{wang2018visual} & 86.2 & 97.2 & 99.4 & 85.3 & 72.4 & 74.0 & 85.7\\
				CAPLS \scriptsize{\cite{wang2019unifying}} & 90.6 & 98.6 & 99.6 & 88.6 & \underline{75.4} & \underline{76.3} & {88.2}\\
				\hline
				SPL (Ours) & 92.7 & \underline{98.7} & \underline{99.8} & \underline{93.0} & \textbf{76.4} & \textbf{76.8} & \textbf{89.6} \\
				\hline
			\end{tabular}%
		}
	}
\end{table}

\begin{table}[!t]
	\centering
	{
		\centering
		\caption[]{Classification Accuracy (\%) on ImageCLEF-DA dataset using either ResNet50 features or ResNet50 based deep models.
		}
		\label{table:uda_clef}
		\resizebox{1\columnwidth}{!}{%
			\begin{tabular}{cccccccc}
				\hline
				Method &I$\to$P & P$\to$I & I$\to$C & C$\to$I & C$\to$P & P$\to$C & Avg \\ \hline
				RTN\cite{long2016unsupervised} & 75.6 & 86.8 & 95.3 & 86.9 & 72.7 & 92.2 & 84.9\\
				MADA\cite{pei2018multi} & 75.0 & 87.9 & 96.0 & 88.8 & 75.2 & 92.2 & 85.8 \\
				iCAN\cite{zhang2018collaborative} & 79.5 & 89.7 & 94.7 & 89.9 & 78.5 & 92.0 & 87.4 \\
				CDAN-E\cite{long2018conditional} & 77.7 & 90.7 & \textbf{97.7} & 91.3 & 74.2 & 94.3 & 87.7\\
				SymNets\cite{zhang2019domain} & \textbf{80.2} & \underline{93.6} & \underline{97.0} & \underline{93.4} & \underline{78.7} & \textbf{96.4} & \underline{89.9}\\
				\hline
				MEDA\cite{wang2018visual} & \underline{79.7} & 92.5 & 95.7 & 92.2 & 78.5 & 95.5 & 89.0\\
				\hline
				SPL (Ours) & 78.3 & \textbf{94.5} & 96.7 & \textbf{95.7} & \textbf{80.5} & \underline{96.3} & \textbf{90.3}\\
				\hline
			\end{tabular}%
		}
	}
\end{table}

In this section, we describe our experiments on four commonly used domain adaptation datasets (i.e. Office+Caltech \cite{gong2012geodesic}, Office31 \cite{saenko2010adapting}, ImageCLEF-DA \cite{caputo2014imageclef} and Office-Home \cite{venkateswara2017deep}). Our approach is firstly compared with state-of-the-art UDA approaches to evaluate its effectiveness. An ablation study is conducted to demonstrate the effects of different components and hyper-parameters in our approach. Finally, we investigate how different hyper-parameters affect the performance.

\subsection{Datasets}\label{sec:dataset}
\textbf{Office+Caltech} \cite{gong2012geodesic} consists of four domains: Amazon (A, images downloaded from online merchants), Webcam (W, low-resolution images by a web camera), DSLR (D, high-resolution images by a digital SLR camera) and Caltech-256 (C).  Ten common classes from all four domains are used: backpack, bike, calculator, headphone, computer-keyboard, laptop-101, computer-monitor, computer-mouse, coffee-mug, and video-projector. There are 2533 images in total with 8 to 151 images per category per domain.
\textbf{Office31} \cite{saenko2010adapting} consists of three domains: Amazon (A), Webcam (W) and DSLR (D). There are 31 common classes for all three domains containing 4,110 images in total. 
\textbf{ImageCLEF-DA} \cite{caputo2014imageclef} consists of four domains. We follow the existing works \cite{zhang2019domain} using three of them in our experiments: Caltech-256 (C), ImageNet ILSVRC 2012 (I), and Pascal VOC 2012 (P). There are 12 classes and 50 images for each class in each domain. 
\textbf{Office-Home} \cite{venkateswara2017deep} is another dataset recently released for evaluation of domain adaptation algorithms. It consists of four different domains: Artistic images (A), Clipart (C), Product images (P) and Real-World images (R). There are 65 object classes in each domain with a total number of 15,588 images.

\subsection{Experimental Setting}\label{sec:setting}
The algorithm is implemented in Matlab\footnote{Code is available: https://github.com/hellowangqian/domain-adaptation-capls}. We use the deep features commonly used in existing works for a fair comparison with the state of the arts. As a result, the Decaf6 \cite{donahue2014decaf} features (activations of the 6\textit{th} fully connected layer of a convolutional neural network trained on ImageNet, $d=4096$) are used for the Office-Caltech dataset. For the other three datasets, ResNet50 \cite{he2016deep} has been commonly used to extract features or as the backbone of deep models in the literature, hence we use ResNet50 features ($d=2048$) in our experiments.  Although a small dimensionality of the PCA space $d_1$ is preferred for less computation, it can cause information loss for a dataset with a large number of classes (e.g., Office-Home). To trade off, we set the values of $d_1$ based on the number of classes in the dataset which results in $d_1=128,512,128$ and $1024$ for Office-Caltech, Office-31, ImageCLEF-DA and Office-Home respectively. For the dimensionality of the space learned by SLPP, we set $d_2=128$ uniformly for all datasets. The number of iterations $T$ is set to 10 in all experiments unless otherwise specified.

\begin{table*}[!htbp]
	\centering
	{
		\centering
		\caption[]{Classification Accuracy (\%) on Office-Home dataset using either ResNet50 features or ResNet50 based deep models.}
		\label{table:uda_o65}
		\resizebox{2\columnwidth}{!}{%
			\begin{tabular}{cccccccccccccc}
				\hline
				Method & A$\to$C & A$\to$P & A$\to$R & C$\to$A&C$\to$P & C$\to$R&P$\to$A & P$\to$C & P$\to$R & R$\to$A & R$\to$C & R$\to$P & Average \\ \hline
				JAN\cite{long2017deep} & 45.9 & 61.2 & 68.9 & 50.4 & 59.7 & 61.0 & 45.8 & 43.4 & 70.3 & 63.9 & 52.4 & 76.8 & 58.3\\
				CDAN-E \cite{long2018conditional} & 50.7 & 70.6 & 76.0 & 57.6 & 70.0 & 70.0 & 57.4 & 50.9 & 77.3 & 70.9 & 56.7 & 81.6 & 65.8\\
				SymNets \cite{zhang2019domain} & 47.7 & 72.9 & 78.5 & 64.2 & 71.3 & 74.2 & 64.2 & 48.8 & 79.5 & \textbf{74.5} & 52.6 & 82.7 & 67.6 \\ 
				TADA \cite{wang2019transferable} & 53.1 & 72.3 & 77.2 & 59.1 & 71.2 & 72.1 & 59.7 & \underline{53.1} & 78.4 & \underline{72.4} & \textbf{60.0} & 82.9 & 67.6 \\
				\hline
				MEDA\cite{wang2018visual} & \underline{54.6} & 75.2 & 77.0 & 56.5 & 72.8 & 72.3 & 59.0 & 51.9 & 78.2 & 67.7 & \underline{57.2} & 81.8 & 67.0\\
				CAPLS \cite{wang2019unifying} & \textbf{56.2} & \textbf{78.3} & \underline{80.2} & \textbf{66.0} & \underline{75.4} & \underline{78.4} & \textbf{66.4} & \textbf{53.2} & \underline{81.1} & {71.6} & 56.1 & \underline{84.3} & \underline{70.6}\\
				\hline
				SPL (Ours) & 54.5 & \underline{77.8} & \textbf{81.9} & \underline{65.1} & \textbf{78.0} & \textbf{81.1} & \underline{66.0} & \underline{53.1} & \textbf{82.8} & 69.9 & 55.3 & \textbf{86.0} & \textbf{71.0} \\
				\hline
			\end{tabular}%
		}
	}
\end{table*}
\subsection{Comparison with State-of-the-Art Approaches}\label{sec:exp_uda}
We compare our approach with the state of the arts including those based on deep features (extracted using deep models such as ResNet50 pre-trained on ImageNet) and deep learning models. The classification accuracy of our approaches and the comparative ones are shown in Tables \ref{table:uda_o10}-\ref{table:uda_o65} in terms of each combination of ``source" $\rightarrow$ ``target" domains and the average accuracy over all different combinations. In each table, the results of deep learning based models are listed on the top followed by deep feature based methods including ours. Our approach is denoted as SPL (Selective Pseudo-Labeling). We use bold and underlined fonts to indicate the best and the second best results respectively in each setting.

From Tables \ref{table:uda_o10}-\ref{table:uda_o65} we can see that our proposed approach with the combination of two different pseudo-labeling methods achieves the highest average accuracy (see the last column of each table) consistently on four datasets. Specifically, our proposed SPL achieves an average accuracy of 93.0\% on the Office-Caltech dataset (Table \ref{table:uda_o10}), slightly better than MEDA \cite{wang2018visual} which has an average accuracy of 92.8\%. On the Office31 dataset (Table \ref{table:uda_o31}), SPL achieves the best or the second-best performance in five out of six tasks and the best average performance of 89.6\% while the second-highest average accuracy (88.4\%) was achieved by the deep learning based approaches SymNets \cite{zhang2019domain} and \cite{wang2019transferable}. On the ImageCLEF-DA dataset (Table \ref{table:uda_clef}), the proposed SPL approach performs the best or the second-best in four out of six tasks and ranks the first with the average accuracy of 90.5\% followed by the deep learning model SymNets \cite[89.9\%]{zhang2019domain} and deep feature based model MEDA  \cite[89.0\%]{wang2018visual}. On the Office-Home dataset (Table \ref{table:uda_o65}, again, our approach SPL outperforms all state-of-the-art models with an average accuracy of 71.0\% against 70.6\% by CAPLS \cite{wang2019unifying} and 67.6\% by SymNets \cite{zhang2019domain} and TADA \cite{wang2019transferable}.

In summary, our selective pseudo-labeling approach can outperform both deep learning models and traditional feature transformation approaches on four commonly used datasets for UDA.

\begin{figure*}
	\centering
	{\includegraphics[width=\textwidth]{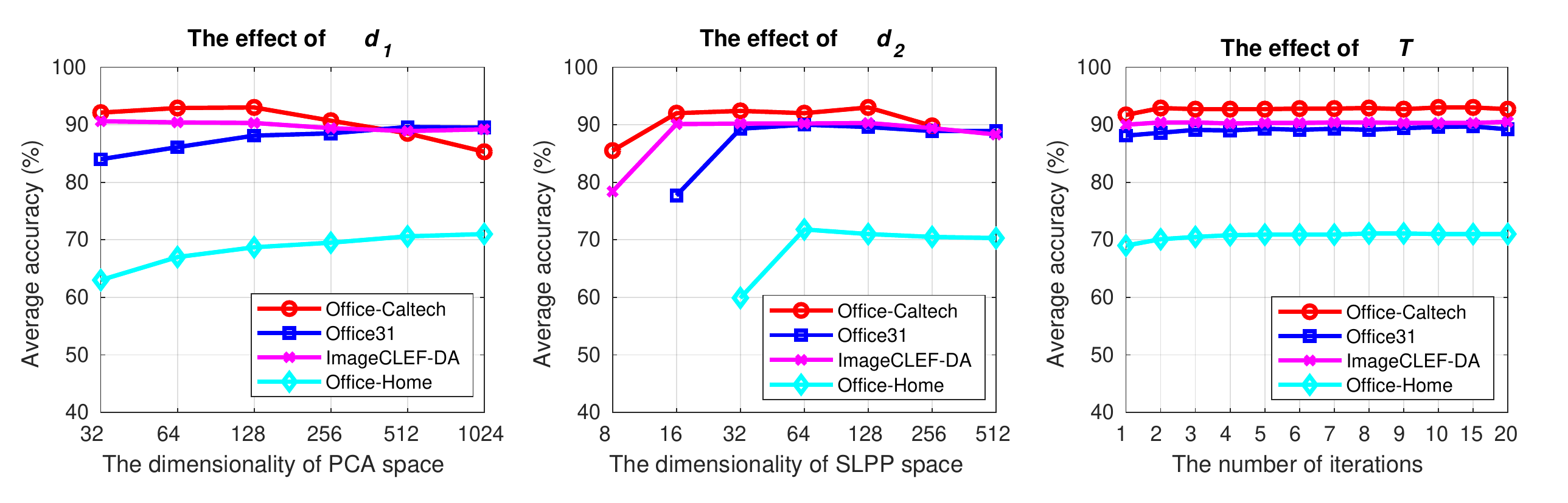}}
	{\caption{The effect of hyper-parameters (i.e. the dimensionality of PCA space $d_1$, the dimensionality of SLPP sapce $d_2$ and the number of iterations $T$).}
		\label{fig:hyper-parameters}}
\end{figure*}

\subsection{Ablation Study}
We conduct an ablation study to analyse how different components of our work contribute to the final performance. To this end, we investigate different combinations of four components: pseudo-labeling (PL), sample selection (S) for pseudo-labeling, nearest class prototype (NCP) and structured prediction (SP). We report the average classification accuracy on four datasets in Table \ref{table:ablation}. It can be observed that methods with pseudo-labeling outperform those without pseudo-labeling and the use of selective pseudo-labeling further improves the performance significantly on all datasets. In terms of the pseudo-labeling strategy, structured prediction (SP) outperforms nearest class prototype (NCP) consistently while the combination of these two can further improve the accuracy marginally.
\begin{table}[!t]
	\centering
	{
		\centering
		\caption[]{Results of ablation study.}
		\label{table:ablation}
		\resizebox{.95\columnwidth}{!}{%
			\begin{tabular}{c|c|c|c|cccc}
				\hline
				\multicolumn{4}{c|}{Method} & \multirow{2}{*}{\footnotesize{Office-Caltech}} & \multirow{2}{*}{\footnotesize{Office31}} & \multirow{2}{*}{\footnotesize{ImageCLEF-DA}} & \multirow{2}{*}{\footnotesize{Office-Home}} \\ \cline{1-4}
				PL& S & NCP & SP & & & & \\ \hline
				\xmark & \xmark & \cmark & \xmark & 81.8 & 82.0 & 86.2 & 63.9 \\
				\xmark & \xmark & \xmark & \cmark & 90.3 & 87.5 & 89.5 & 68.0\\
				\xmark & \xmark & \cmark & \cmark & 90.7 & 87.6 & 89.4 & 68.1\\ \hline
				
				\cmark & \xmark & \cmark & \xmark & 85.5 & 83.7 & 86.9 & 66.2\\
				\cmark & \xmark & \xmark & \cmark & 91.9 & 88.0 & 90.0 & 68.9\\
				\cmark & \xmark & \cmark & \cmark & 92.0 & 88.0 & 90.0 & 69.0 \\
				\hline
				\cmark & \cmark & \cmark & \xmark & 90.8 & 87.8 & 89.0 & 70.8\\
				\cmark & \cmark & \xmark & \cmark & 93.0 & 89.5 & 90.2 & 71.0\\
				\cmark & \cmark & \cmark & \cmark & 93.0 & 89.6 & 90.3 & 71.0\\
				\hline
			\end{tabular}%
		}
	}
\end{table}

\subsection{Effects of Hyper-parameters}
Our approach has three hyper-parameters: the dimensionality of PCA space $d_1$, the dimensionality of SLPP space $d_2$ and the number of iterations $T$. We investigate how each hyper-parameter affects the performance by setting it to a series of different values while fixing the other two. The results are shown in Figure \ref{fig:hyper-parameters} in which the average accuracy over all possible source-target pairs are reported for four datasets. As we can see, the datasets with more classes (e.g., Office31 and Office-Home) tend to large values of $d_1$ while small values of $d_1$ are beneficial to the datasets with fewer classes (e.g., Office-Caltech and ImageCLEF-DA). In terms of the dimensionality of SLPP space $d_2$, the performance does not change  too much unless its value is less than the number of classes. The number of iterations $T$ has nearly zero effect on the performance when it is greater than 2. To summarize, our approach is not sensitive to the hyper-parameters and performs comparably well if only $d_2$ is set greater than the number of classes.
\section{Conclusion}\label{sec:conclusion}
We propose a novel selective pseudo-labeling approach to UDA by incorporating supervised subspace learning and structured prediction based pseudo-labeling into an iterative learning framework. The proposed approach outperforms other state-of-the-art methods on four benchmark datasets. The ablation study demonstrates the effectiveness of selective pseudo-labeling and structured prediction which can also be employed to train the deep learning models for UDA in the future work.

\bibliographystyle{aaai}
\footnotesize
\bibliography{ref}

\newpage
\appendix
\section{Supplementary Material}
\begin{figure*}[htpb]
	\centering
	{\includegraphics[width =\textwidth]{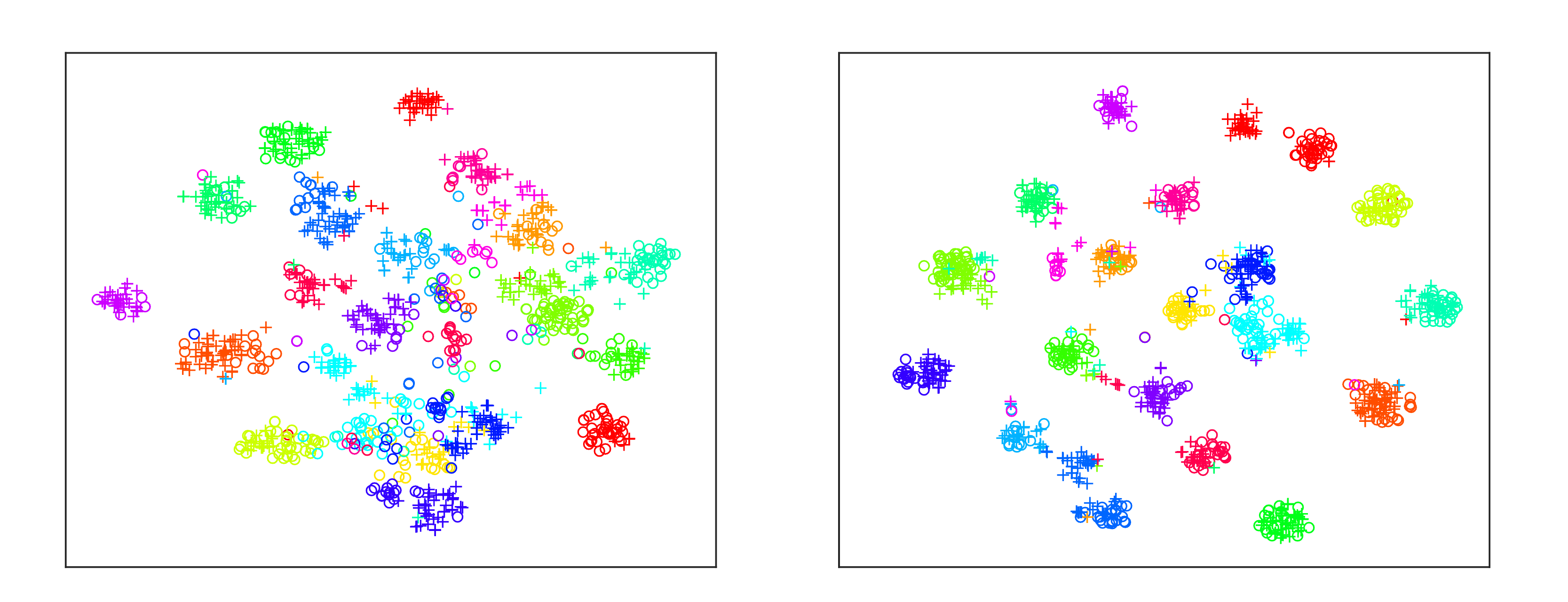}}
	{\caption{Data distributions in the original deep feature space (left) and the learned subspace (right) for 20 classes from the Home-Office dataset. ``o" and ``+" denote the source and target samples respectively. Different classes are denoted by different colors.}
		\label{fig:tsne}}
\end{figure*}
\begin{figure*}[t!]
	\centering
	{\includegraphics[width =\textwidth]{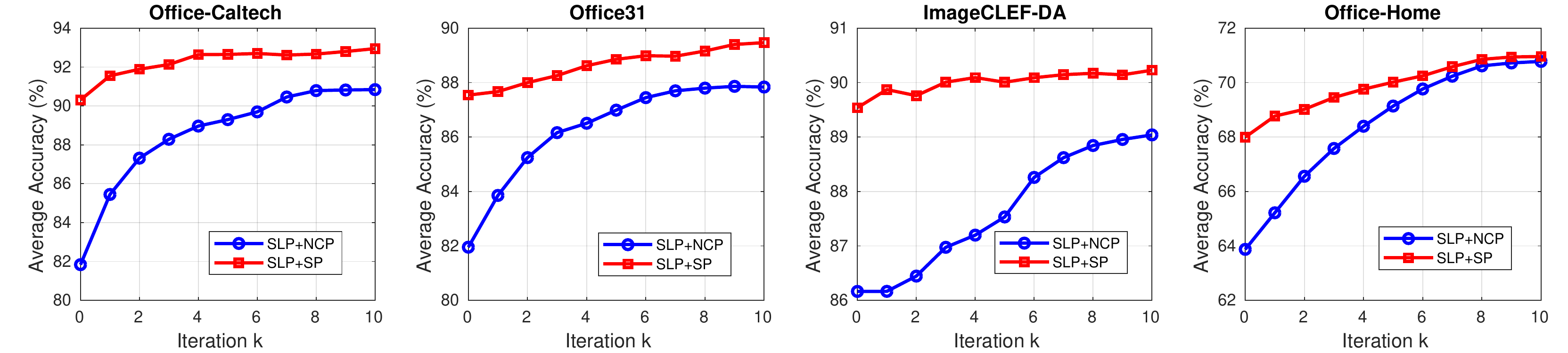}}
	{\caption{Accuracy curves of iterative learning process using nearest class prototype (NCP) and structured prediction (SP).}
		\label{fig:learning_curve}}
\end{figure*}
Due to the page limit, we present more experimental results and discussion in this supplementary material.

\subsection{Visualisation of Domain Alignment}\label{sec:introduction}
To provide intuitive insights of domain alignment results using the proposed algorithm, we take the most challenging Office-Home dataset \cite{venkateswara2017deep} as an example to visualise the data distributions before and after domain alignment. We randomly select 20 out of 65 classes from this dataset for a clear view. The t-SNE \cite{maaten2008visualizing} technique is used to enable the 2D visualisation. Figure \ref{fig:tsne} shows the visualisation results in the original deep feature space (left, before the domain alignment) and the learned subspace (right, after the domain alignment). The symbol ``o" is used to denote samples from the source domain whilst ``+" is for samples from the target domain. Different classes are indicated by different colors.
 
It can be seen that our domain alignment algorithm is able to promote the separability in the learned subspace (right). Both the marginal and conditional distributions are well aligned. It can also be observed from Figure \ref{fig:tsne} that data distributions in the original deep feature space are already good enough due to the transferrability of deep models pre-trained on ImageNet \cite{deng2009imagenet}. This is also validated by the results (Table \ref{table:baseline}) of two baseline methods, i.e., 1 Nearest Neighbor (1NN) and Support Vector Machine (SVM), based solely on the labeled source samples without any domain adaptation.

It is such favorable properties of deep features that enables the structured prediction algorithm to be able to take advantage of the cluster information and promote the domain alignment performance. 

\begin{table}[h]
	\centering
	{
		\centering
		\caption[]{Results of 1NN and SVM without domain adaptation.
		}
		\label{table:baseline}
		\resizebox{.95\columnwidth}{!}{%
			\begin{tabular}{ccccc}
				\hline
				Method & Office-Caltech & Office31 & ImageCLEF-DA & Office-Home \\ \hline
				1NN &83.8 &79.8 &80.0 & 54.5\\
				SVM &83.9 &81.0 &85.9 & 61.3\\ 
				\hline
			\end{tabular}%
		}
	}
\end{table}

\subsection{Learning Process of NCP and SP}
We have demonstrated the superiority of the Structured Prediction (SP) over the Nearest Class Prototype (NCP) method in the ablation study. Here we show more details of the learning process when two different methods are employed in Figure \ref{fig:learning_curve}. In this experiment, we set the number of iterations to 10 and report the classification accuracy for each iteration. 

It can be seen from Figure \ref{fig:learning_curve} that with more iterations (more pseudo-labeled target samples), the classification performance is improved gradually. The SP method can achieve much better performance than the NCP method at the beginning of the iterative learning process. It is due to the benefit of exploiting the structural information in the target domain. With more iterations when more target samples are pseudo-labeled and participate in the domain alignment, the classification accuracy can be improved gradually for both SP and NCP methods, although SP can always achieve better final performance than NCP.

\subsection{Response to Reviewers}
\subsubsection{Reviewer1}
\textit{While the results on Office-Home dataset are very good, a number of benchmarks datasets such as CIFAR, SVHN were omitted in the comparison. Can the authors explain why a more versatile collection of datasets was not used ? in that regards, what is the main limitation of the method ?}

The selected four datasets in our experiments have been widely used in recent publications. We were aware of other datasets used for UDA including those mentioned by the reviewer (e.g., MNIST-SVHN, CIFAR-STL, etc.). We select the datasets containing more domains so that there are more sub-tasks (i.e., three domains lead to six sub-tasks and four domains lead to 12 sub-tasks) on which the proposed method can be thoroughly evaluated and compared against state-of-the-arts fairly. We have also conducted experiments on the CIFAR$\rightarrow$STL and STL$\rightarrow$CIFAR tasks using ResNet50 features and achieved 94.9\% and 82.4\% accuracy respectively. However, our approach is not suitable for the Digit datasets due to the limitation of excessive memory usage when the number of samples is too large. This limitation is shared by many feature transformation based methods and addressing it can be an interesting direction for our future work. 

\subsubsection{Reviewer2}

\textit{I've seen the computational complexity section, but can you comment on how longs do the experiments take in practice?}

Running on a laptop with Intel i5-7300 CPU and 32G RAM, it takes around 18, 83, 10, 2070 seconds for all sub-tasks of the Office-Caltech, Office31, ImageCLEF-DA and Office-Home datasets, respectively.

\subsubsection{Reviewer3}
\textit{Q1. In the second paragraph p.3, the authors stated "the structured prediction based method tends to select samples far away from the source samples". I cannot find the reason in the paper.}

Thanks for pointing out this ambiguous statement in the manuscript. We would like to give an explanation here and revise the statement in the manuscript accordingly to dismiss the potential ambiguity and misleading. Most existing sample selection strategies select the target samples close to the source class prototypes (i.e. source class means) for pseudo labeling. As a result, the selected target samples are usually the outliers of the corresponding target cluster. One can imagine two point clusters (which are not completely overlapped due to the domain shift) representing the source sample cluster and the corresponding target sample cluster belonging to the same class, then the closest target point to the source cluster center must locate in the outermost shell of the target point cluster. In contrast, our structured prediction based method aims to select the target samples closest to the "target" cluster center rather than the "source" cluster center for pseudo labeling. As a result, the selected target samples are not necessarily the "closest" ones to the source samples and this is why we state "the structured prediction based method tends to select samples far away from the source samples".
The pseudo-labeled target samples will participate the next iteration of supervised learning and therefore will be "pulled" to the source data of the same class. If the pseudo-labeled target samples are only the outliers or the outermost ones of the target cluster, the result of the "pulling" will only force the source and target clusters to "touch" each other. However, if the pseudo-labeled target samples are close to the target cluster center, the result of the "pulling" will force the two clusters align. This explains why our structured prediction based method enables a faster (i.e. less iterations) domain alignment.

\textit{Q2. Why the pseudo-labeling approach is employed? It seems not to be a straightforward implementation of the cluster assumption. Cannot the assumption be implemented as a loss function like as, for example, a similar formulation to the semi-supervised loss, which encourages the decision boundary not to cross the dense region of unsupervised instances?}

We agree that a straightforward implementation of the cluster assumption is to learn decision boundaries passing the low-density region of unlabeled samples. Although it works well in semi-supervised learning where the unlabeled samples come from the same distribution as the labeled samples, it is not necessarily a good solution to the domain adaptation problem. The reason is, in the domain adaptation problem, the unlabeled target samples and labeled source samples are from different domains and hence they can be distributed in separate clusters even they belong to the same class. As a result, the decision boundaries learned under the cluster assumption for UDA may not be class discriminative (they might be boundaries of domains rather than classes, see Figure 1 (a) in [2]). Actually, the existing works[1][2] following the cluster assumption for UDA also employ pseudo-labeling to address this issue.
As we have discussed in the section of related work, pseudo-labeling is one of the enabling techniques applied in many unsupervised domain adaptation approaches towards the alignment of both marginal and conditional distributions. Our approach aims to learn a subspace where the two domains are aligned via supervised LPP which requires pseudo-labeled samples from the target domain for conditional distribution alignment. The assumption of cluster structure in the target domain is explored by our structured prediction algorithm which predicts the pseudo-labels of target samples collectively (i.e. cluster-wisely). In this sense, the cluster assumption has been used in a totally different way from learning decision boundaries passing the low-density region.
In summary, the cluster assumption in UDA is different from that in the semi-supervised learning problem; and we explore the cluster structural information in the target domain via our structured prediction algorithm rather than the low-density separation idea for semi-supervised learning.

[1] Shu, Rui, et al. "A dirt-t approach to unsupervised domain adaptation." ICLR 2018.

[2] Kumar, Abhishek, et al. "Co-regularized alignment for unsupervised domain adaptation." NIPs 2018.
%

\end{document}